\documentclass[11pt,a4paper]{article}
\usepackage[hyperref]{acl2021}
\usepackage{times}
\usepackage{latexsym}

\usepackage{microtype}

\usepackage{subfig}
\usepackage{graphicx}

\aclfinalcopy 
\def\aclpaperid{810} 

\usepackage{amsmath}
\usepackage{multirow}
\usepackage{amsfonts}
\newcommand{\tabincell}[2]{\begin{tabular}{@{}#1@{}}#2\end{tabular}}  
\newcommand{\ub}[1]{\underline{\textbf{#1}}}

\newcommand{\An}[1]{\textcolor{black}{#1}}

\title{Learning Algebraic Recombination for Compositional Generalization}

\author{
  Chenyao Liu$^{1}$\thanks{\quad Work done during an internship at Microsoft Research. The first two authors contributed equally to this paper.} \quad Shengnan An$^{2*}$ \quad Zeqi Lin$^{3}$\thanks{\quad Corresponding author.} \quad Qian Liu$^{4*}$ \quad Bei Chen$^3$ \\
  \textbf{ \quad Jian-Guang LOU$^3$  \quad Lijie Wen$^{1\dagger}$ \quad Nanning Zheng $^2$ \quad Dongmei Zhang$^3$}\\
  
  $^1$ School of Software, Tsinghua University \quad $^2$ Xi'an Jiaotong University\\
  $^3$ Microsoft Research Asia \quad $^4$ Beihang University\\
  \texttt{\{liucy19@mails, wenlj@\}.tsinghua.edu.cn} \\
  \texttt{\{an1006634493@stu, nnzheng@mail\}.xjtu.edu.cn} \\
  \texttt{\{Zeqi.Lin, beichen, jlou, dongmeiz\}@microsoft.com} \\
  \texttt{qian.liu@buaa.edu.cn} 
}

\date{}

\begin{document}
\maketitle
\begin{abstract}
Neural sequence models exhibit limited \emph{compositional generalization} ability in semantic parsing tasks. Compositional generalization requires \emph{algebraic recombination}, i.e., dynamically recombining structured expressions in a recursive manner.
However, most previous studies mainly concentrate on recombining lexical units, which is an important but not sufficient part of algebraic recombination.
In this paper, we propose \textsc{LeAR}, an end-to-end neural model to learn algebraic recombination for compositional generalization.
The key insight is to model the semantic parsing task as a homomorphism between a latent syntactic algebra and a semantic algebra, thus encouraging algebraic recombination.
Specifically, we learn two modules jointly: a Composer for producing latent syntax, and an Interpreter for assigning semantic operations.
Experiments on two realistic and comprehensive compositional generalization benchmarks demonstrate the effectiveness of our model.
The source code is publicly available at \href{https://github.com/microsoft/ContextualSP}{https://github.com/microsoft/ContextualSP}.
\end{abstract}

\section{Introduction}

\begin{figure}
  \centering
  \subfloat[][Compositional generalization requires \textbf{algebraic recombination}, i.e., dynamically recombining structured expressions in a recursive manner.\label{fig:prod}]{\includegraphics[width=0.37\textwidth]{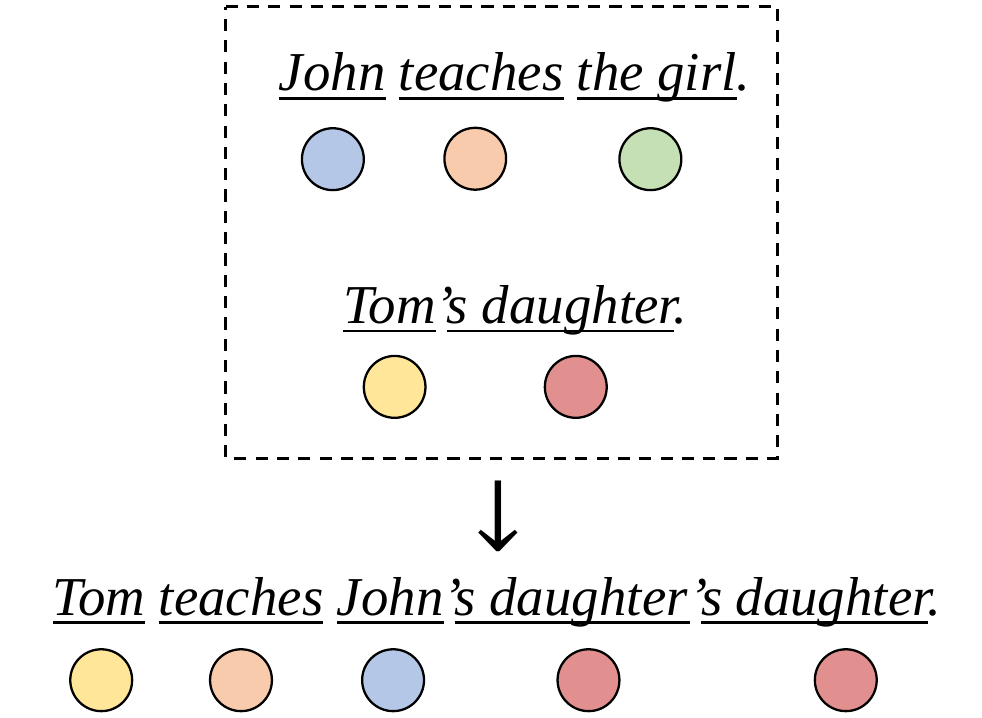}}\\
  \subfloat[][Most previous studies mainly concentrate on recombining lexical units, which is an important but not sufficient part of algebraic recombination.\label{fig:sys}]{\includegraphics[width=0.45\textwidth]{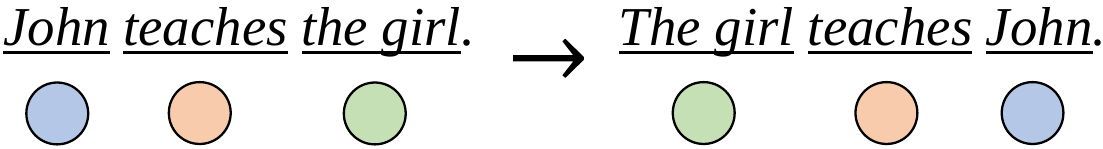}}
  \caption{Compositional generalization.
  }
\end{figure}

The principle of compositionality is an essential property of language: the meaning of a complex expression is fully determined by its structure and the meanings of its constituents \cite{pelletier2003context-tpoc, szabo2004compositionality-tpoc}.
Based on this principle, human intelligence exhibits \textit{compositional generalization} --- the algebraic capability to understand and produce a potentially infinite number of novel expressions by dynamically recombining known components \cite{chomsky1957syntactic-hicg, fodor_connectionism_1988-hicg, fodor2002compositionality-hicg}.
For example, people who know the meaning of ``\textit{John teaches the girl}'' and ``\textit{Tom's daughter}'' must know the meaning of
``\textit{Tom teaches John's daughter's daughter}'' (Figure \ref{fig:prod}), even though they have never seen such complex sentences before.

In recent years, there has been accumulating evidence that end-to-end deep learning models lack such ability in semantic parsing (i.e., translating natural language expressions to machine interpretable semantic meanings) tasks \cite{lake2018scan, keysers2019measuring, kim2020cogs, tsarkov2020cfq-lcg}.

Compositional generalization requires \textbf{algebraic recombination}, i.e., dynamically recombining structured expressions in a recursive manner.
In the example in Figure \ref{fig:prod}, understanding ``\emph{John's daughter's daughter}'' is a prerequisite for understanding ``\emph{Tom teaches John's daughter's daughter}'', while ``\emph{John's daughter's daughter}'' is also a novel compound expression, which requires recombining ``\emph{John}'' and ``\emph{Tom's daughter}'' recursively.

Most previous studies on compositional generalization mainly concentrate on recombining lexical units (e.g., words and phrases) \cite{nips2019dataaug,li2019compositional,andreas2019dataaug,Gordon2020spstru,aky2020dataaug,guo2020dataaugl,russin2019spstru}, of which an example is shown in Figure \ref{fig:sys}.
This is a necessary part of algebraic recombination, but it is not sufficient for compositional generalization.
There have been some studies on algebraic recombination \cite{liu2020spstru, chen2020spstru}.
However, they are highly specific to a relative simple domain SCAN \cite{lake2018scan} and can hardly generalize to more complex domains.

In this paper, our main point to achieve algebraic recombination is to \textbf{model semantic parsing as a homomorphism between a latent syntactic algebra and a semantic algebra} \cite{montague1970universal-algre, marcus2019algebraic}.
Based on this formalism, we focus on learning the high-level mapping between latent syntactic operations and semantic operations, rather than the direct mapping between expression instances and semantic meanings.

Motivated by this idea, we propose \textsc{LeAR} (\textbf{Le}arning \textbf{A}lgebraic \textbf{R}ecombination), an end-to-end neural architecture for compositional generalization.
\textsc{LeAR} consists of two modules: a \textit{Composer} and an \textit{Interpreter}.
Composer learns to model the latent syntactic algebra, thus it can produce the latent syntactic structure of each expression in a bottom-up manner;
Interpreter learns to assign semantic operations to syntactic operations, thus we can transform a syntactic tree to the final composed semantic meaning.



Experiments on two realistic and comprehensive compositional generalization benchmarks (\textit{CFQ} \cite{keysers2019measuring} and \textit{COGS} \cite{kim2020cogs}) demonstrate the effectiveness of our model:
CFQ $67.3\%\to 90.9\%$, COGS $35.0\%\to 97.7\%$.


\section{Compositionality: An Algebraic View}\label{section:algebra}

A semantic parsing task aims to learn a meaning-assignment function $m: L\to M$, where $L$ is the set of (simple and complex) expressions in the language, and $M$ is the set of available semantic meanings for the expressions in $L$.
Many end-to-end deep learning models are built upon this simple and direct formalism, in which the principle of compositionality is not leveraged, thus exhibiting limited compositional generalization.

To address this problem, in this section we put forward the formal statement that ``\textit{compositionality requires the existence of a homomorphism between the expressions of a language and the meanings of those expressions}'' \cite{montague1970universal-algre}.

Let us consider a language as a partial algebra $\mathbf{L} = \langle L, (f_\gamma)_{\gamma\in\Gamma}\rangle$, where $\Gamma$ is the set of underlying syntactic (grammar) rules, and we use $f_\gamma: L^k\to L$ to denote the syntactic operation with a fixed arity $k$ for each $\gamma\in\Gamma$.
Note that $f_\gamma$ is a partial function, which means that we allow $f_\gamma$ be undefined for certain expressions.
Therefore, $\mathbf{L}$ is a partial algebra, and we call it a \textbf{syntactic algebra}.
In a semantic parsing task, $\mathbf{L}$ is latent, and we need to model it by learning from data.

Consider now $\mathbf{M} = \langle M, G\rangle$, where $G$ are semantic operations upon $M$.
$\textbf{M}$ is also a partial algebra, and we call it a \textbf{semantic algebra}.
In a semantic parsing task, we can easily define this algebra (by enumerating all available semantic primitives and semantic operations), since $\mathbf{M}$ is a machine-interpretable formal system.

The key to compositionality is that the meaning-assignment function $m$ should be a homomorphism from $\mathbf{L}$ to $\mathbf{M}$.
That is, for each $k$-ary syntactic operation $f_\gamma$ in $\mathbf{L}$,
there exists a $k$-ary semantic operation $g_\gamma\in G$ such that whenever $f_\gamma(e_1, ..., e_k)$ is defined,
\begin{equation}\label{eq:homo}
m(f_\gamma(e_1, ..., e_k)) = g_\gamma(m(e_1), ..., m(e_k)).
\end{equation}

Based on this formal statement, the task of learning the meaning-assignment function $m$ can be transformed as two sub-tasks: (1) learning latent syntax of expressions (i.e., modeling the syntactic algebra $\mathbf{L}$); (2) learning the operation assignment function $(f_\gamma)_{\gamma\in\Gamma}\to G$.

\noindent \textbf{Learning latent syntax}.
We need to learn a syntactic parser that can produce the syntactic structure of each given expression.
To ensure compositional generalization, there must be an underlying grammar (i.e., $\Gamma$), and we hypothesize that $\Gamma$ is a context-free grammar.

\noindent \textbf{Learning operation assignment}.
In the syntax tree, for each nonterminal node with $k$ nonterminal children, we assign a $k$-ary semantic operation to it.
This operation assignment entirely depends on the underlying syntactic operation $\gamma$ of this node.

In semantic parsing tasks, we do not have respective supervision for these two sub-tasks.
Therefore, we need to jointly learning these two sub-tasks only from the end-to-end supervision $\mathcal{D}\subset L\times M$.

\section{Model}\label{sec:Model}

\begin{figure}[t]
    \centering
    \includegraphics[width=0.5\textwidth]{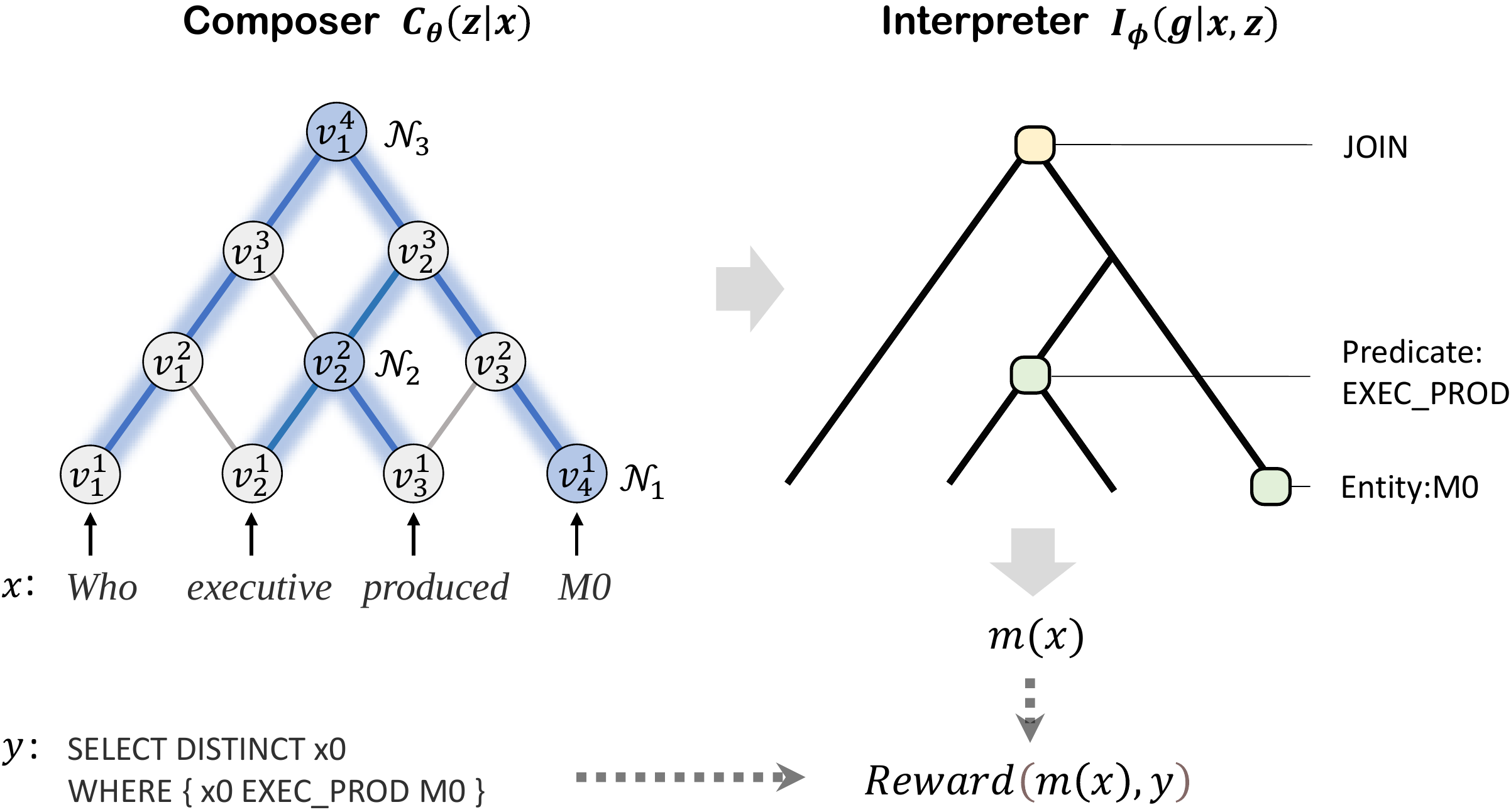}
    \caption{An overview of \textsc{LeAR}: (1) Composer $C_\theta(z|x)$ is a neural network based on latent Tree-LSTM, which produces the latent syntax tree $z$ of input expression $x$; (2) Interpreter $I_\phi(g|x, z)$ is a neural network that assigns a semantic operation for each nonterminal node in $z$.}
    \label{fig:framework}
\end{figure}

We propose a novel end-to-end neural model \textsc{LeAR} (\textbf{Le}arning \textbf{A}lgebraic \textbf{R}ecombination) for compositional generalization in semantic parsing tasks.
Figure \ref{fig:framework} shows its overall architecture.
\textsc{LeAR} consists of two parts:
(1) \emph{Composer} $C_\theta(z|x)$, which produces the latent syntax tree $z$ of input expression $x$;
(2) Interpreter $I_\phi(g|x, z)$, which assigns a semantic operation for each nonterminal node in $z$.
$\theta$ and $\phi$ refers to learnable parameters in them respectively.
We generate a semantic meaning $m(x)$ according to the predicted $z$ and $g$ in a symbolic manner, then check whether it is semantic equivalent to the ground truth semantic meaning $y$ to produce rewards for optimizing $\theta$ and $\phi$.

\subsection{Composer}

We use $x=[x_1,...,x_T]$ to denote an input expression of length $T$.
Composer $C_\theta(z|x)$ will produce a latent binary tree $z$ given $x$.


\subsubsection{Latent Tree-LSTM}

We build up the latent binary tree $z$ in a bottom-up manner based on Tree-LSTM encoder, called latent Tree-LSTM \cite{choi2018learning-lt, havrylov2019cooperative-lt}.

Given the input sequence $x$ of length $T$, latent Tree-LSTM merges two nodes into one parent node at each merge step, constructing a binary tree after $T-1$ merge steps.
The merge process is implemented by selecting the adjacent node pair which has the highest merging score.

At the $t$-th ($1\leq t < T$) merge step, we have:
\begin{equation}\label{eq:ihat}
\hat{i}_t = \mathop{\arg\max}_{1\leq i \leq T-t}\text{Linear} (\text{Tree-LSTM} (  \mathbf{r}_i^t, \mathbf{r}_{i+1}^t ) )
\end{equation}

Here ``Tree-LSTM'' is the standard child-sum tree-structured LSTM encoder \cite{tree_lstm_2015-trls}.
We use $v_i^t$ to denote the $i$-th cell at layer $t$ (the $t$-th merge step is determined by the $t$-th layer), and use $\mathbf{r}_i^t$ to denote the representation of $v_i^t$:
\begin{equation}
    \mathbf{r}_{i}^{1}
    =\operatorname{Linear}(\operatorname{Emb}(x_i))
\end{equation}
\begin{equation}\label{eq:treelstm}
\mathop{\mathbf{r}_{i}^{t}}_{t>1}=
\begin{cases}
\mathbf{r}_{i}^{t-1} & i<\hat{i}_{t-1}\\
\operatorname{Tree-LSTM} (  \mathbf{r}_i^{t-1}, \mathbf{r}_{i+1}^{t-1} ) & i=\hat{i}_{t-1}\\
\mathbf{r}_{i+1}^{t-1} & i>\hat{i}_{t-1}\\
\end{cases}
\end{equation}

Then we can obtain a unlabeled binary tree, in which $\{v_1^1, v_2^1, ..., v_T^1\}$ are leaf nodes, and $\{v_{\hat{i}_1}^2, v_{\hat{i}_2}^3\, ..., v_{\hat{i}_{T-1}}^{T}\}$ are non-leaf nodes.





\subsubsection{Abstraction by Nonterminal Symbols}\label{sec:REDUCE}

As discussed in Section \ref{section:algebra}, our hypothesis is that the underlying grammar $\Gamma$ is context-free.
Therefore, each syntactic rule $\gamma\in\Gamma$ can be expressed in the form of:
\begin{equation}
\nonumber
A\to B,\qquad A\in\mathcal{N}, B\in(\mathcal{N} \cup \Sigma)^+
\end{equation}
where $\mathcal{N}$ is a finite set of nonterminals, and $\Sigma$ is a finite set of terminal symbols.

\textbf{Abstraction} is an essential property of context-free grammar:
each compound expression $e$ will be abstracted as a simple nonterminal symbol $\mathcal{N}(e)$, then it can be combined with other expressions to produce more complex expressions, no matter what details $e$ originally has.
This setup may benefit the generalizability, thus we want to incorporate it as an inductive bias into our model.


Concretely, we assume that there are at most $N$ latent nonterminals in language $\mathbf{L}$ (i.e., $\mathcal{N}=\{\mathcal{N}_1,..., \mathcal{N}_N\}$, where $N$ is a hyper-parameter).
For each node $v_i^t$ in tree $z$, we perform a $(N+1)$-class classification:
\begin{equation}
\hat{c}_{v_i^t} = \mathop{\arg\max}_{0\leq c\leq N}\operatorname{Linear}(\mathbf{r}_i^t)\\
\end{equation}

We assign the nonterminal $\mathcal{N}_{\hat{c}_{v_i^t}}$ to $v_i^t$ when $\hat{c}_{v_i^t}>0$.
The collection of such nonterminal nodes are denoted as $V'_{z}$.
Then we modify Equation \ref{eq:treelstm}:
\begin{equation}\label{eq:r}
\begin{split}
\mathop{\mathbf{r}_{i}^{t}}_{t>1} &=
\begin{cases}
\mathbf{r}_{i}^{t-1} & i<\hat{i}_{t-1}\\
\operatorname{Tree-LSTM} (  \overline{\mathbf{r}}_i^{t-1}, \overline{\mathbf{r}}_{i+1}^{t-1} ) & i=\hat{i}_{t-1}\\
\mathbf{r}_{i+1}^{t-1} & i>\hat{i}_{t-1}\\
\end{cases} \\
\overline{\mathbf{r}}_{i}^{t} &=
\begin{cases}
\operatorname{Linear}(\operatorname{Emb}(\mathcal{N}(v_i^t))) & v_i^t\in V'_{z}\\
\mathbf{r}_{i}^{t} & v_i^t\not\in V'_{z}\\
\end{cases}
\end{split}
\end{equation}

Equation \ref{eq:r} means that: in nonterminal nodes, the bottom-up message passing will be reduced from $\mathbf{r}_i^t$ to a nonterminal symbol $\mathcal{N}(v_i^t)$, thus mimicking the abstraction setup in context-free grammar.

\subsection{Interpreter}

For each nonterminal node $v\in V'_{z}$, Interpreter $I_\phi(g|x, z)$ assigns a semantic operation $g_v$ to it.

We divide nonterminal nodes into two categories:
(1) \textit{lexical nodes}, which refer to those containing no any other nonterminal node in the corresponding sub-trees;
(2) \textit{algebraic nodes}, which refer to the rest of nonterminal nodes.

\paragraph{Interpreting Lexical Nodes}

For each lexical node $v$, Interpreter assigns a semantic primitive (i.e., $0$-ary semantic operation) to it.
Take the CFQ benchmark as an example:
it uses SPARQL queries to annotate semantic meanings, thus semantic primitives in CFQ are entities (e.g., \textit{m.0gwm\_wy}), predicates (e.g., \textit{ns:film.director.film}) and attributes (e.g., \textit{ns:people.person.gender m\_05zppz}).

We use a classifier to predict the semantic primitive:
\begin{equation}\label{eq:classifier}
g_v = \mathop{\arg\max}_{g\in G_{lex}}\operatorname{Linear}(\mathbf{h}_{v, x})
\end{equation}
where $G_{lex}$ is the collection of semantic primitives in the domain, and $\mathbf{h}_{v, x}$ is the contextual representation of the span corresponding to $v$ (implemented using Bi-LSTM).
\textit{Contextually conditioned variation} is an important phenomenon in language: the meaning of lexical units varies according to the contexts in which they appear \cite{allwood2003meaning-ccv}.
For example, ``\emph{editor}'' means a predicate ``\emph{film.editor.film}'' in expression ``\emph{Is M0 an editor of M1?}'', while it means an attribute ``\emph{film.editor}'' in expression ``\emph{Is M0 an Italian editor?}''.
This is the reason why we use contextual representation in Equation \ref{eq:classifier}.

\paragraph{Interpreting Algebraic Nodes}

For each algebraic node $v$, Interpreter assigns a semantic operation to it.
The collection of all possible semantic operations $G_{opr}$ also depends on the domain.
Take the CFQ benchmark as an example\footnote{It is not difficult to define $G_{lex}$ and $G_{opr}$ for each domain, as semantic meanings are always machine-interpretable. The semantic operations of another compositional generalization benchmark, COGS, are listed in the Appendix.}, this domain has two operations (detailed in Table \ref{tab:CFQ_operations}): $\wedge$ (conjunction) and JOIN.

We also use a classifier to predict the semantic operation of $v$:
\begin{equation}\label{eq:opr}
g_v = \mathop{\arg\max}_{g\in G_{opr}}\operatorname{Linear}(\mathbf{r}_v)
\end{equation}
where $\mathbf{r}_v$ is the latent Tree-LSTM representation of node $v$ (see Equation \ref{eq:r}).

In Equation \ref{eq:opr}, we do not use any contextual information from outside $v$.
This setup is based on the assumption of \textbf{semantic locality}: each compound expression should mean the same thing in different contexts.

\begin{table}[t]
    \renewcommand\arraystretch{1.1}
    \centering
    \scalebox{0.79}{
    \begin{tabular}{ccl}
    \hline
    \textbf{Operation} & \begin{tabular}[c]{@{}c@{}}\small{\textbf{Args[$t_1$, $t_2$]}}$\rightarrow$\\\small{\textbf{Result Type}}\end{tabular} & \multicolumn{1}{c}{\textbf{Example}} \\ \hline
    \multirow{7}{*}{$\wedge$($t_1$, $t_2$)} 
    & [P, P]$\rightarrow$P & Who [\ub{direct} and \ub{act}] M0? \\ \cline{2-3} 
     & [E, E]$\rightarrow$E & Who direct [\ub{M0} and \ub{M1}]? \\ \cline{2-3} 
     & [A, A]$\rightarrow$A & Is M0 an [\ub{Italian} \ub{female}]? \\ \cline{2-3} 
     & [A, E]$\rightarrow$E & \multirow{2}{*}{Is [\ub{M0} an \ub{Italian female}]?} \\
     & [E, A]$\rightarrow$E &  \\ \cline{2-3} 
     & [A, P]$\rightarrow$P & \multirow{2}{*}{Is M0 M3's [\ub{Italian} \ub{editor}]?} \\
     & [P, A]$\rightarrow$P &  \\ \hline
    \multirow{4}{*}{JOIN($t_1$, $t_2$)} 
    & [E, P]$\rightarrow$E & \multirow{2}{*}{Is M0 an [\ub{editor} of \ub{M1}]?} \\
     & [P, E]$\rightarrow$E &  \\ \cline{2-3} 
     & [A, P]$\rightarrow$E & \multirow{2}{*}{Who [\ub{marries} an \ub{Italian}]?} \\
     & [P, A]$\rightarrow$E &  \\ \hline
    \end{tabular}
    }
    \caption{Semantic operations in CFQ.
    A/P/E represents Attribute/Predicate/Entity.}
    \label{tab:CFQ_operations}
\end{table}




\section{Training}


Denote $\tau = \{z, g\}$ as the trajectory produced by our model where $z$ and $g$ are actions produced from  Composer and Interpreter, respectively, and $R(\tau)$ as the reward of trajectory $\tau$ (elaborated in Sec.~\ref{sec:reward}).
Using policy gradient \cite{policy_gradient_1999} with the likelihood ratio trick, our model can be optimized by ascending the following gradient:
\begin{equation}\label{eq:gradient}
    \nabla \mathcal{J}(\theta, \phi)=\mathbb{E}_{\tau \sim {\pi}_{\theta, \phi}}\;R(\tau)\nabla \log \pi_{ \theta, \phi}\left(\tau\right),
\end{equation}
where $\theta$ and $\phi$ are learnable parameters in Composer and Interpreter respectively and $\nabla$ is the abbreviation of $\nabla_{ \theta, \phi}$. 
Furthermore, the REINFORCE algorithm \cite{reinforce_algo_1992} is leveraged to approximate Eq.~\ref{eq:gradient} and the mean-reward baseline \cite{reinforce_baseline_2001} is employed to reduce variance.

\subsection{Reward Design}\label{sec:reward}
The reward $R\left({\tau}\right)$ combines two parts as:
\begin{equation}
    R\left({\tau}\right) = 
    \alpha \cdot R_1\left({\tau}\right) + (1-\alpha) \cdot R_2\left({\tau}\right),
\end{equation}

\noindent \textbf{Logic-based Reward} $R_1(\tau)$.
We use $m(x$) and $y$ to denote the predicted semantic meaning and the ground truth semantic meaning respectively.
Each semantic meaning can be converted to a conjunctive normal form\footnote{
For example, the semantic meaning of ``\emph{Who directed and edited $M_0$ 's prequel and $M_1$?}'' can be converted to a conjunctive normal form with four components:
``$x_0\cdot\text{DIRECT}\cdot x_1\cdot\text{PREQUEL}\cdot M_0$'',
``$x_0\cdot\text{EDIT}\cdot x_1\cdot\text{PREQUEL}\cdot M_0$'',
``$x_0\cdot\text{DIRECT}\cdot M_1$'', and
``$x_0\cdot\text{EDIT}\cdot M_1$''.}.
We use $S_{m(x)}$ and $S_y$ to denote conjunctive components in $m(x$) and $y$, then define $R_1(\tau)$ based on Jaccard similarity (i.e., intersection over union):
\begin{equation}\label{eq:reward sentence CFQ}
    R_1\left({\tau}\right) = 
    \text{Jaccard-Sim}(S_{m(x)}, S_y)
\end{equation}

\noindent \textbf{Primitive-Based Reward} $R_2(\tau)$.
We use $S'_{m(x)}$ and $S'_y$ to denote semantic primitives ocurred in $m(x$) and $y$.
Then we define $R_2(\tau)$ as:
\begin{equation}\label{eq:reward_atom}
    R_{2}\left({\tau}\right) = 
    \text{Jaccard-Sim}(S'_{m(x)}, S'_y)
\end{equation}

\subsection{Reducing Search Space}
To reduce the huge search space of $\tau$, we make two constraints as follows.

\noindent \textbf{Parameter Constraint}.
Consider $v$, a tree node with $n (n>0)$ nonterminal children.
Composer will never make $v$ a nonterminal node, if no semantic operation has $n$ parameters.

\noindent \textbf{Phrase Table Constraint.}
Following the strategy proposed in \citet{guo2020hierarchical}, we build a ``phrase table'' consisting of lexical units (i.e., words and phrases) paired with semantic primitives that frequently co-occur with them\footnote{Mainly based on statistical word alignment technique in machine translation, detailed in the Appendix.}.
Composer will never produce a lexical node outside of this table, and Interpreter will use this table to restrict candidates in Equation \ref{eq:classifier}.


\subsection{Curriculum Learning} \label{sec:curriculum}

To help the model converge better, we use a simple curriculum learning \cite{bengio2009curriculum} strategy to train the model.
Specifically, we first train the model on samples of input length less than a cut-off $N_{CL}$, then further train it on the full train set.




\section{Experimental Setup}

\begin{figure}
\begin{minipage}[tp]{\linewidth}
\begin{minipage}[b]{\linewidth}
\centering
  \includegraphics[width=0.9\textwidth]{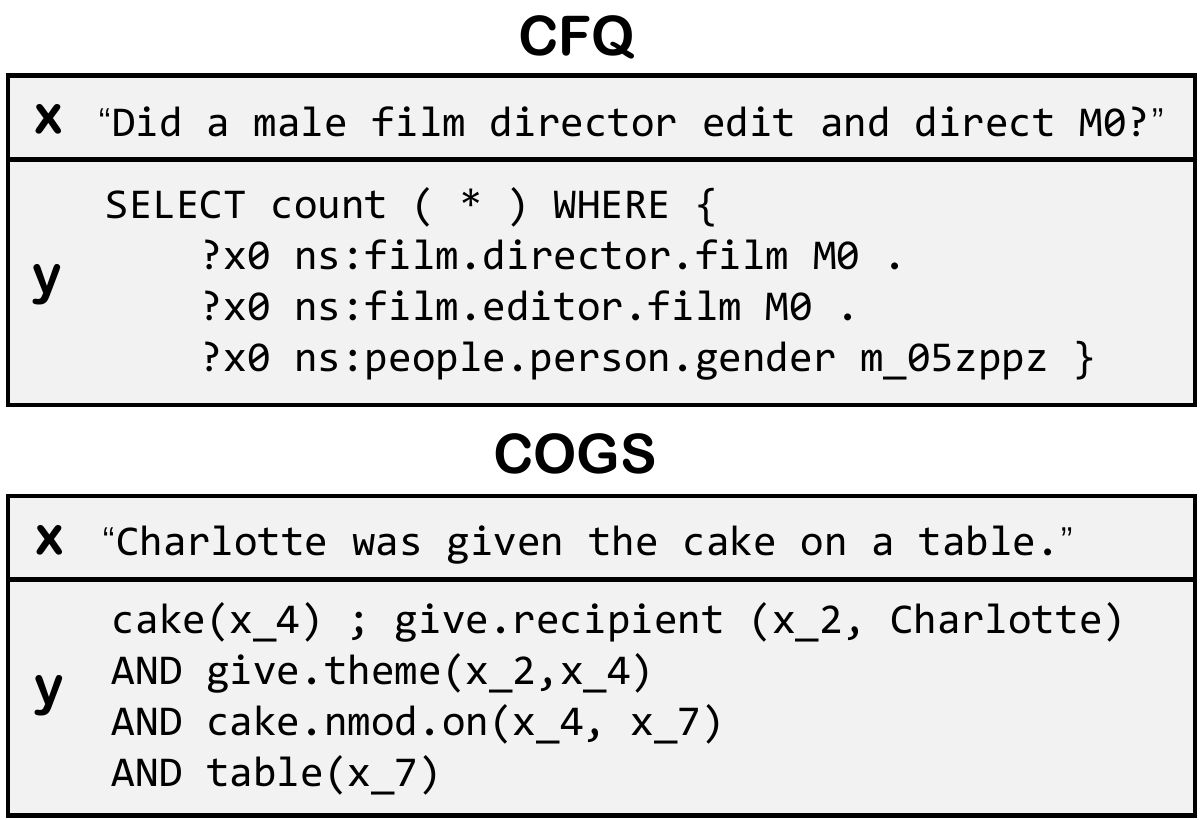}
  \captionof{figure}{Examples of CFQ and COGS.}
  \label{fig:example}
\end{minipage}

\vspace{0.3cm}

\begin{minipage}[b]{\linewidth}
\centering
\resizebox{.9\linewidth}{!}{
\begin{tabular}{ccc}
\hline
\textbf{Statistics} & \textbf{CFQ} & \textbf{COGS}   \\
\hline
Train Size & 95,743 & 24,155\\
Dev Size & 11,968 & 3,000\\
Test Size & 11,968 & 21,000\\
Vocab Size & 96 & 740	\\
\hline
Avg Input Len (Train/Test)  & 13.5/15.1 & 7.5/9.8 \\
Avg Output Len (Train/Test)  & 27.7/34.0  & 43.6/67.6 \\
Input Pattern Coverage\footnote{Input/output pattern coverage is the percentage of test $x$/$y$ whose patterns occur in the train data. Output patterns are determined by anonymizing semantic primitives, and input patterns are determined by anonymizing their lexical units.} & 0.022 & 0.783 \\
Output Pattern Coverage & 0.045 & 0.782 \\
\hline
\end{tabular}}
\captionof{table}{Dataset statistics.}
\label{tab:dataset}
\end{minipage}
 \end{minipage}
 \end{figure}

\noindent \textbf{Benchmarks}.
We mainly evaluate \textsc{LeAR} on \textit{CFQ} \cite{keysers2019measuring} and \textit{COGS} \cite{kim2020cogs}, two comprehensive and realistic benchmarks for measuring compositional generalization.
They use different semantic formulations:
CFQ uses SPARQL queries, and COGS uses logical queries (Figure \ref{fig:example} shows examples of them).
We list dataset statistics in Table \ref{tab:dataset}.
The input/output pattern coverage indicates that:
CFQ mainly measures the algebraic recombination ability, while COGS measures both lexical recombination ($\sim78\%$) and algebraic recombination ($\sim22\%$).

In addition to these two compositional generalization benchmarks in which utterances are synthesized by formal grammars, we also evaluate \textsc{LeAR} on \textit{GEO} \cite{zelle1996learning-csp}, a widely used semantic parsing benchmark, to see whether \textsc{LeAR} can generalize to utterances written by real users.
We use the variable-free FunQL \cite{kate2005geofunql} as the semantic formalism, and we follow the compositional train/test split \cite{finegan2018geosplit} to evaluate compositional generalization.


\begin{table*}
\centering
\resizebox{.9\linewidth}{!}{
\begin{tabular}{lcccc}
\hline \textbf{Models} & \textbf{MCD-MEAN} & \textbf{MCD1} & \textbf{MCD2} & \textbf{MCD3} \\ \hline
LSTM+Attention \cite{keysers2019measuring} & 14.9$\pm$1.1 &	28.9$\pm$1.8 &	5.0$\pm$0.8 &	10.8$\pm$0.6 \\
Transformer \cite{keysers2019measuring} & 17.9$\pm$0.9 &	34.9$\pm$1.1 &	8.2$\pm$0.3 &	10.6$\pm$1.1 \\
Universal Transformer \cite{keysers2019measuring} & 18.9$\pm$1.4 &	37.4$\pm$2.2 &	8.1$\pm$1.6 &	11.3$\pm$0.3 \\
Evolved Transformer \cite{furrer2020compositional}  & 	20.8$\pm$0.7 &	42.4$\pm$1.0 &	9.3$\pm$0.8 &	10.8$\pm$0.2 \\
\hline 
T5-11B \cite{furrer2020compositional} & 40.9$\pm$4.3	& 61.4$\pm$4.8 &	30.1$\pm$2.2 &	31.2$\pm$5.7 \\
T5-11B-mod \cite{furrer2020compositional} & 42.1$\pm$9.1 &	61.6$\pm$12.4 &	31.3$\pm$12.8 &	33.3$\pm$2.3 \\
\hline 
Neural Shuffle Exchange \cite{furrer2020compositional} & 2.8$ \pm $0.3 &	5.1$\pm$0.4 &	0.9$\pm$0.1 &	2.3$\pm$0.3 \\
CGPS \cite{furrer2020compositional, li2019compositional}  & 7.1$\pm$1.8 &	13.2$\pm$3.9 &	1.6$\pm$0.8 &	6.6$\pm$0.6 \\
HPD \cite{guo2020hierarchical} & 67.3$\pm$4.1 & 72.0$\pm$7.5 &	66.1$\pm$6.4 &	63.9$\pm$5.7 \\
\hline 
\textbf{\textsc{LeAR}} & \textbf{90.9$\pm$1.2} & \textbf{91.7$\pm$1.0} & \textbf{89.2$\pm$1.9} & \textbf{91.7$\pm$0.6} \\
\hspace{2em}  w/o Abstraction & 85.4$\pm$4.5 & 88.4$\pm$1.6 & 80.0$\pm$11 & 87.9$\pm$0.8 \\
\hspace{2em}  w/o Semantic locality & 87.9$\pm$2.7 & 89.8$\pm$1.7 & 87.3$\pm$1.8 & 86.5$\pm$4.6 \\
\hspace{2em}  w/o Primitive-based reward & 85.3$\pm$7.8 & 77.0$\pm$19 & 89.2$\pm$2.2 & 89.7$\pm$2.1 \\
\hspace{2em}  w/o Curriculum learning & 71.9$\pm$15.4 & 59.7$\pm$23 & 77.2$\pm$13.5 & 78.8$\pm$9.6 \\
\hspace{2em}  w/o Tree-LSTM & 30.4$\pm$3.2 & 40.1$\pm$1.9 & 25.6$\pm$6.1 & 25.4$\pm$1.8 \\
\hline
\end{tabular}}
\caption{Accuracy on three splits (MCD1/MCD2/MCD3) of CFQ benchmark. }
\label{tab:CFQ_res}
\end{table*}

\noindent \textbf{Baselines}.
For CFQ, we consider 3 groups of models as our baselines:
(1) sequence-to-sequence models based on deep encoder-decoder architecture, including LSTM+Attention \cite{hochreiter1997long, bahdanau2014neural}, Transformer \cite{vaswani2017attention}, Universal Transformer \cite{dehghani2018universal} and Evolved Transformer \cite{so2019evolved};
(2) deep models with large pretrained encoder, such as T5 \cite{raffel2019exploring};
(3) Models that are specially designed for compositional generalization, which include Neural Shuffle Exchange Network \cite{freivalds2019neural}, CGPS \cite{li2019compositional}, and state-of-the-art model HPD \cite{guo2020hierarchical}.
For COGS, we quote the baseline results in the original paper \cite{kim2020cogs}.
For GEO, we take the baseline results reported by \citet{herzig2020span-csp}, and also compare with two specially designed methods: SpanBasedSP \cite{herzig2020span-csp} and PDE \cite{guo2020geoiterative}.

\noindent \textbf{Evaluation Metric}.
We use accuracy as the evaluation metric, i.e., the percentage test samples of which the predicted semantic meaning $m(x)$ is semantically equivalent to the ground truth $y$.

\noindent \textbf{Hyper-Parameters}.
We set $N=3/2/3$ (the number of nonterminal symbols), and $\alpha=0.5/1.0/0.9$ for CFQ/COGS/GEO respectively.
In CFQ, the curriculum cut-off $N_{CL}$ is set to 11, as we statistically find that this is the smallest curriculum that contains the complete vocabulary.
We do not apply curriculum learning strategy to COGS and GEO, as \textsc{LeAR} can work well without curriculum learning in both benchmarks.
Learnable parameters ($\theta$ and $\phi$) are optimized with AdaDelta \cite{adadelta_2012-apd}, and the setting of learning rate is discussed in Section \ref{section:ablation}.
We take the model that performs best on the validation set for testing, and all results are obtained by averaging over 5 runs with different random seeds.
See Appendix for more implementation details.

\section{Results and Discussion}

\begin{table}[t]
    \centering
    \resizebox{\linewidth}{!}{
    \begin{tabular}{lc}
    \hline
    \textbf{Model} & \textbf{Acc} \\ \hline
    Transformer \cite{kim2020cogs} & 35 ± 6 \\
    LSTM (Bi) \cite{kim2020cogs}  & 16 ± 8 \\
    LSTM (Uni) \cite{kim2020cogs}  & 32 ± 6 \\ \hline
    \textbf{\textsc{LeAR}} & \textbf{97.7 ± 0.7} \\ 
    \hspace{2em} w/o Abstraction & 94.5 ± 2.8 \\ 
    \hspace{2em}  w/o Semantic locality & 94.0 ± 3.6 \\ 
    \hspace{2em}  w/o Tree-LSTM & 80.7 ± 4.3 \\
    \hline
    \end{tabular}}
    \caption{Accuracy on COGS benchmark. }
    \label{result:COGS}
\end{table}

\begin{table}[t]
    \centering
    \resizebox{\linewidth}{!}{
    \begin{tabular}{lc}
    \hline
    \textbf{Model} & \textbf{Acc} \\ \hline
    Seq2Seq \cite{herzig2020span-csp} & 46.0 \\
    BERT2Seq \cite{herzig2020span-csp} & 49.6 \\
    GRAMMAR \cite{herzig2020span-csp} & 54.0 \\ \hline
    PDE \cite{guo2020geoiterative} & 81.2 \\
    SpanBasedSP \cite{herzig2020span-csp} & 82.2 \\
    \textbf{\textsc{LeAR}} & \textbf{84.1} \\ \hline
    \end{tabular}}
    \caption{Accuracy on GEO benchmark. }
    \label{result:GEO}
\end{table}

Table \ref{tab:CFQ_res} shows average accuracy and 95\% confidence intervals on three splits of CFQ.
\textsc{LeAR} achieves an average accuracy of 90.9\% on these three splits, outperforming all baselines by a large margin.
We list some observations as follows.

\noindent \textbf{Methods for lexical recombination cannot generalize to algebraic recombination}.
Many methods for compositional generalization have been proved effective for lexical recombination.
Neural Shuffle Exchange and CGPS are two representatives of them.
However, experimental results show that they cannot generalize to CFQ, which focus on algebraic recombination.

\noindent \textbf{Knowledge of semantics is important for compositional generalization}.
Seq2seq models show poor compositional generalization ability ($\sim20\%$).
Pre-training helps a lot ($\sim20\% \to \sim40\%$), but still not satisfying.
\textsc{HPD} and \textsc{LeAR} incorporate knowledge of semantics (i.e., semantic operations) into the models, rather than simply model semantic meanings as sequences.
This brings large profit.

\noindent \textbf{Exploring latent compositional structure in a bottom-up manner is key to compositional generalization.}
\textsc{HPD} uses LSTM to encode the input expressions, while \textsc{LeAR} uses latent Tree-LSTM, which explicitly explores latent compositional structure of expressions.
This is the key to the large accuracy profit ($67.3\%\to 90.9\%$).

Table \ref{result:COGS} shows the results on COGS benchmark.
It proves that \textsc{LeAR} can well generalize to domains which use different semantic formalisms, by specifying domain-specific $G_{lex}$ (semantic primitives) and $G_{opr}$ (semantic operations).
Table \ref{result:GEO} shows the results on GEO benchmark.
It proves that \textsc{LeAR} can well generalize to utterances written by real users (i.e., non-synthetic utterances).

\subsection{Ablation Study}
\label{section:ablation}

Table \ref{tab:CFQ_res} and \ref{result:COGS} also report results of some ablation models.
Our observations are as follows.

\noindent \textbf{Abstraction by nonterminal symbols brings profit}.
We use ``\textit{w/o abstraction}'' to denote the ablation model in which Equation \ref{eq:r} is disabled.
This ablation leads to $5.5\%/3.2\%$ accuracy drop on CFQ/COGS.

\noindent \textbf{Incorporating semantic locality into the model brings profit}.
We use ``\textit{w/o semantic locality}'' to denote the ablation model in which a Bi-LSTM layer is added before the latent Tree-LSTM.
This ablation leads to $3.0\%/3.7\%$ accuracy drop on CFQ/COGS.

\noindent \textbf{Tree-LSTM contributes significantly to compositional generalization}. 
In the ablation ``\emph{w/o Tree-LSTM}'', we replace the Tree-LSTM encoder with a span-based encoder, in which each span is represented by concatenating its start and end LSTM representations (similar to \citet{herzig2020span-csp}).
In Table \ref{tab:CFQ_res} and \ref{result:COGS}, we can see that span-based encoder severely affects the performance and even much worse than the results of ``\textit{w/o abstraction}'' and ``\textit{w/o semantic locality}''.
This ablation hints that Tree-LSTM is the main inductive bias of compositionality in our model.

\noindent \textbf{Primitive-based reward helps the model converge better}.
The ablation ``\textit{w/o primitive-based reward}'' leads to $5.6\%$ accuracy drop on CFQ, and the model variance has become much larger.
The key insight is: primitive-based reward guides the model to interpret polysemous lexical units more effectively, thus helping the model converge better.

\begin{table}
\centering
\resizebox{\linewidth}{!}{
\begin{tabular}{lcccc}
\hline \textbf{Ratio} & \textbf{MCD-MEAN} & \textbf{MCD1} & \textbf{MCD2} & \textbf{MCD3} \\ \hline
1:1:1  & 87.4$\pm$7.1  & 91.5$\pm$2.1  & \textbf{89.4$\pm$2.3}  &	81.2$\pm$17 \\
1:0.5:0.1  & \textbf{90.9$\pm$1.2} &	\textbf{91.7$\pm$1.0} &	89.2$\pm$1.9 &	\textbf{91.7$\pm$0.6} \\
1:0.1:0.1 & 86.7$\pm$3.9   & 89.4$\pm$1.6 	 & 85.8$\pm$2.7 	 & 84.9$\pm$7.5 	\\
\hline
\end{tabular}}
\caption{Results of different learning rate ratios  of lexical Interpreter, Composer, and algebraic Interpreter.}
\label{tab:lr_setting}
\end{table}

\begin{figure}[tp]
  \centering
  \includegraphics[width=0.48\textwidth]{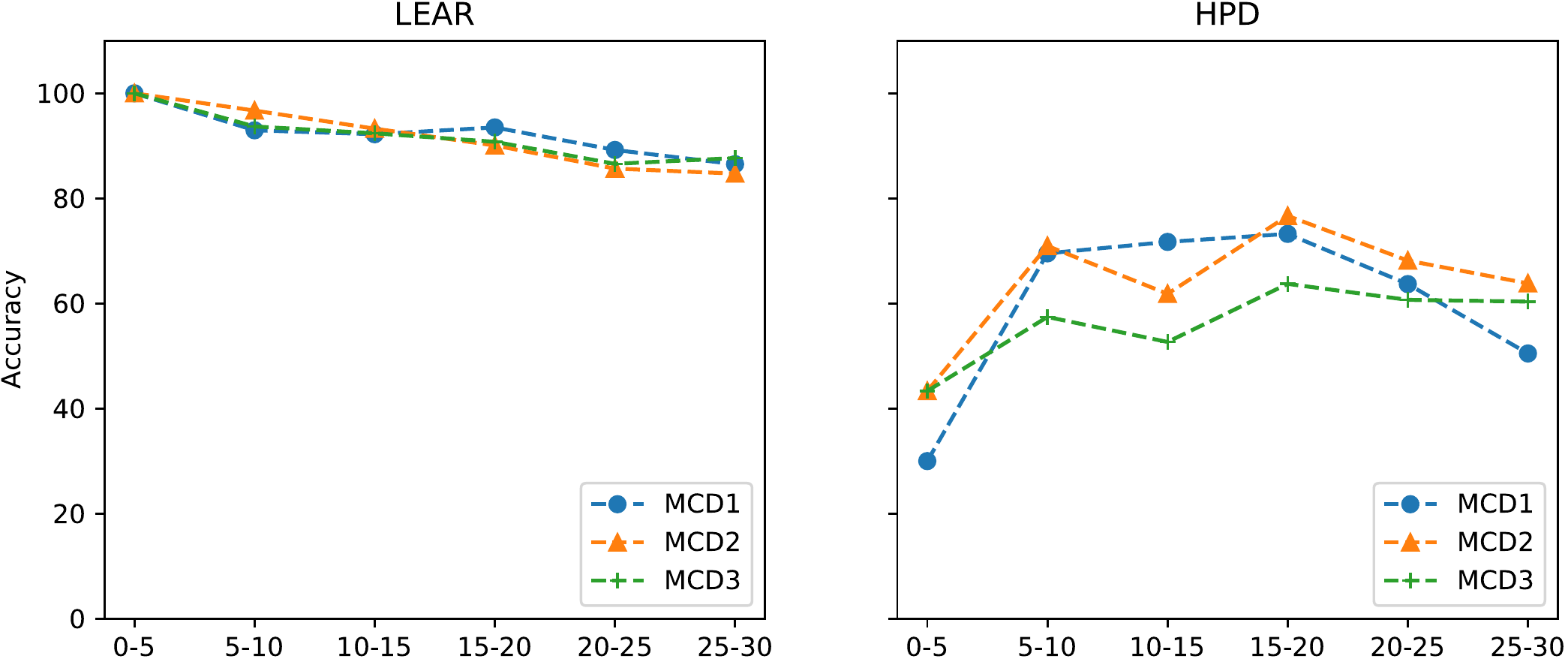}
  \caption{Performance by input length.}
  \label{fig:diff_len}
\end{figure}

\noindent \textbf{Curriculum learning helps the model converge better}.
The ablation ``\emph{w/o curriculum learning}'' leads to 19\% accuracy drop on CFQ, and the model variance has become much larger.
This indicates the importance of curriculum learning.
On COGS, \textsc{LeAR} performs well without curriculum learning.
We speculate that there are two main reasons:
(1) expressions of COGS is much shorter than CFQ;
(2) the input/output pattern coverage of COGS is much higher than CFQ.

\noindent \textbf{Higher component with smaller learning rate}. 
Inspired by the differential update strategy used in \citet{liu2020spstru}(i.e., the higher level the component is positioned in the model, the slower the parameters in it should be updated), we set three different learning rates to three different components in \textsc{LeAR} (in bottom-up order): lexical Interpreter, Composer, and algebraic Interpreter.
We fix the learning rate of lexical Interpreter to 1, and adjust the ratio of the learning rates of Composer and algebraic Interpreter to lexical Interpreter.
Table \ref{tab:lr_setting} shows the results on CFQ.
The hierarchical learning rate setup ($1:0.5:0.1$) achieves the best performance.



\begin{figure}[tp]
  \centering
  
  \subfloat[][Composer error. A correct syntax tree should compose ``\emph{parent of a cinematographer}'' as a constituent, while the predicted syntax tree incorrectly composes ``\emph{a cinematographer played M0}''.\label{fig:type2}]
  {\includegraphics[width=\linewidth]{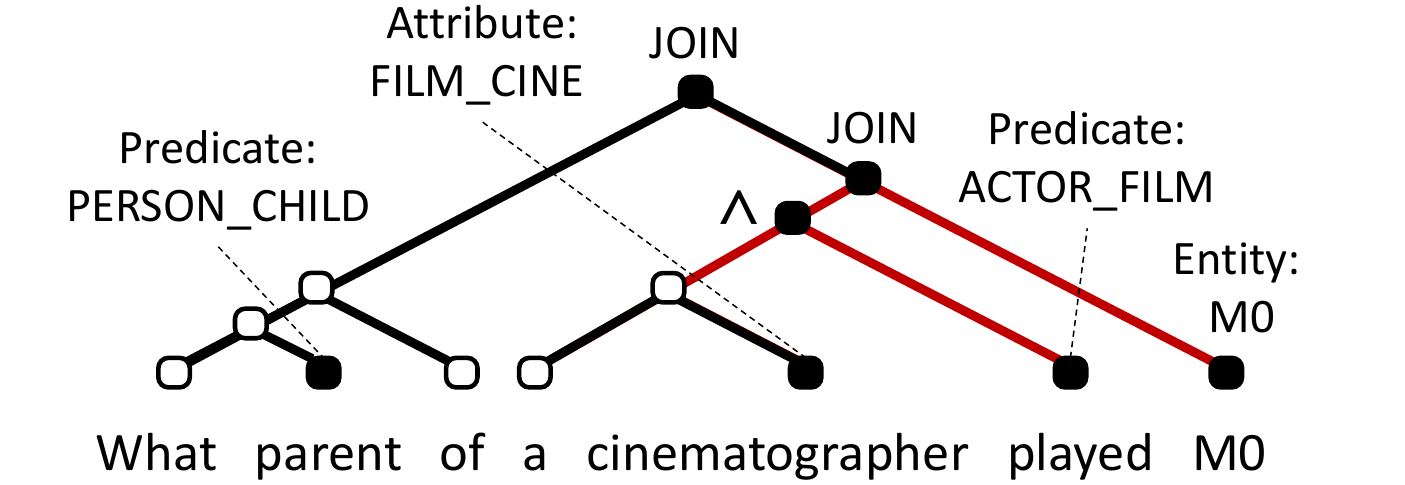}}
  
  \subfloat[][Interpreter error. In this expression, the first ``\emph{influenced}'' should be assigned a semantic primitive ``\emph{influence.influence\_node.influenced}'', while Interpreter incorrectly assigns ``\emph{influence.influence\_node.influenced\_by}'' (abbreviated as ``\emph{INFLU\_BY}'' in this figure) to it.\label{fig:type1}]
  {\includegraphics[width=\linewidth]{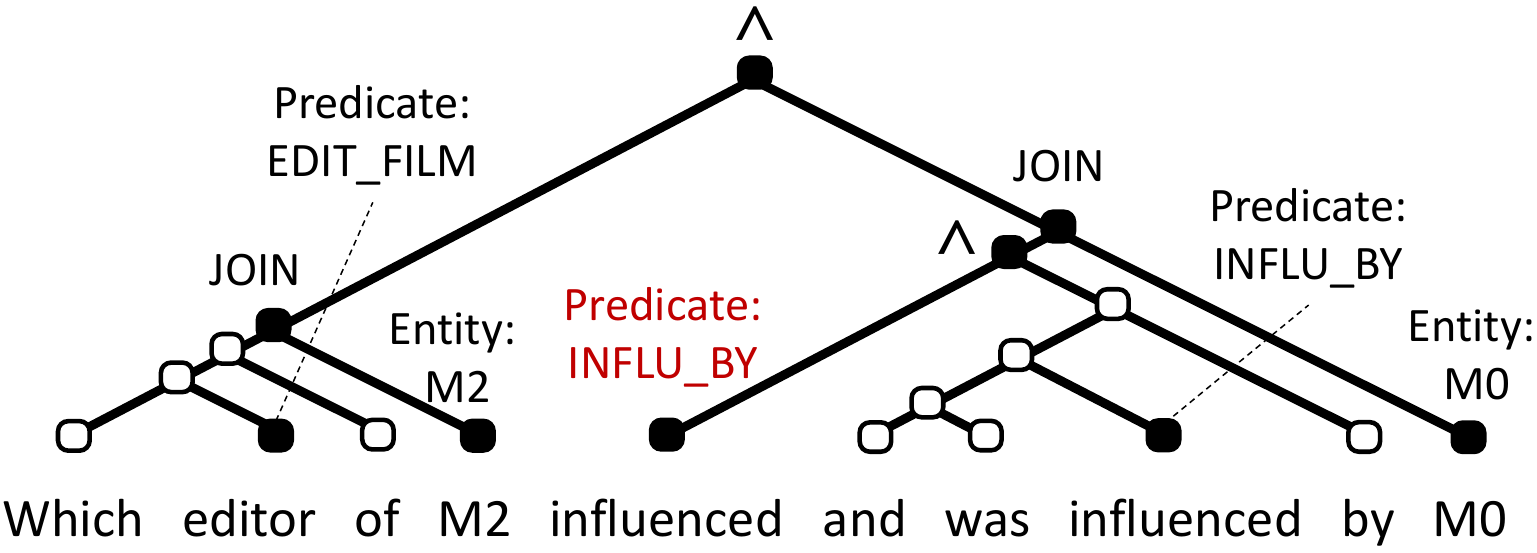}} \\
  
  \caption{Two error cases. We use solid nodes to denote predicted nonterminal nodes. Incorrect parts are colored red.}\label{fig:err_type} 
  
\end{figure}

\subsection{Closer Analysis}

We also conduct closer analysis to the results of \textsc{LeAR} as follows.

\subsubsection{Performance by Input Length}

Intuitively, understanding longer expressions requires stronger algebraic recombination ability than shorter examples.
Therefore, we expect that our model should keep a good and stable performance with the increasing of input length.


Figure \ref{fig:diff_len} shows the performance of \textsc{LeAR} and \textsc{HPD} (the state-of-the-art model on CFQ) under different input lengths.
Specifically, test instances are divided into 6 groups by length: $[1, 5], [6, 10], ..., [26, 30]$), and we report accuracy on each group separately.
The results indicate that \textbf{\textsc{LeAR} has stable high performance for different input lengths}, with only a slow decline as length increases.
Even on the group with the longest input length, \textsc{LeAR} can maintain an average 86.3\% accuracy across three MCD-splits.


\subsubsection{Error Analysis}

To understand the source of errors, we take a closer look at the failed test instances of \textsc{LeAR} on CFQ.
These failed test instances account for $9.1\%$ of the test dataset.
We category them into two error types:

\noindent\textbf{Composer error (CE)}, i.e., test cases where Composer produces incorrect syntactic structures (only considering nonterminal nodes).
Figure \ref{fig:type2} shows an example.
As we do not have ground-truth syntactic structures, we determine whether a failed test instance belongs to this category based on handcraft syntactic templates.

\noindent\textbf{Interpreter error (IE)}, i.e., test cases where Composer produces correct syntactic structures but Interpreter assigns one or more incorrect semantic primitives or operations.
Figure \ref{fig:type1} shows an example, which contains an incorrect semantic primitive assignment.

Table \ref{tab:err_type} shows the distribution of these two error types.
On average, 39.19\% of failed instances are composer errors, and the remaining 60.81\% are interpreter errors.

\begin{table}
\centering
\begin{tabular}{lccc}
\hline \textbf{Error Type} & \textbf{MCD1} & \textbf{MCD2} & \textbf{MCD3} \\ \hline
CE & 45.70\% & 32.05\% & 39.83\% \\
IE & 54.30\% & 67.95\% & 60.17\% \\
\hline
\end{tabular}
\caption{Distribution of CE (Composer Error) and IE (Interpreter Error). }
\label{tab:err_type}
\end{table}



\subsection{Limitations}

Our approach is implicitly build upon the assumption of \textbf{primitive alignment}, that is, each primitive in the meaning representation can align to at least one span in the utterance.
This assumption holds in most cases of various semantic parsing tasks, including CFQ, COGS, and GEO.
However, for robustness and generalizability, we also need to consider cases that do not meet this assumption.
For example, consider this utterance ``\textit{Obama's brother}'', of which the corresponding meaning representation is ``$Slibing(People[Obama])\wedge Gender[Male]$''.
Neither ``$Slibing$'' nor ``$Gender[Male]$'' can align to a span in the utterance, as the composed meaning of them is expressed by a single word (``\textit{brother}'').
Therefore, \textsc{LeAR} is more suitable for formalisms where primitives can better align to natural language.



In addition, while our approach is general for various semantic parsing tasks, the collection of semantic operations needs to be redesigned for each task. We  need to ensure that these semantic operations are $k$-ary projections (as described in Section \ref{section:algebra}), and all the meaning representations are covered by the operations collection.
This is tractable, but still requires some efforts from domain experts.



\section{Related Work}




\subsection{Compositional Generalization}

Recently, exploring compositional generalization (CG) on neural networks has attracted large attention in NLP community.
For SCAN \cite{lake2018scan}, the first benchmark to test CG on seq2seq models, many solutions have been proposed, which can be classified into two tracks:
data augmentation \cite{andreas2019dataaug, aky2020dataaug, guo2020dataaugl} 
and specialized architecture \cite{nips2019dataaug, li2019compositional, Gordon2020spstru}.
However, most of these works only focus on lexical recombination.
Some works on SCAN have stepped towards algebraic recombination \cite{liu2020spstru, chen2020spstru}, but they do not generalize well to other tasks such as CFQ \cite{keysers2019measuring} and COGS \cite{kim2020cogs}.

Before our work, there is no satisfactory solution on CFQ and COGS.
Previous works on CFQ demonstrated that MLM pre-training \cite{furrer2020compositional} and iterative back-translation \cite{guo2020revisiting} can improve traditional seq2seq models.
HPD \cite{guo2020hierarchical}, the state-of-the-art solution before ours, was shown to be effective on CFQ, but still far from satisfactory.
As for COGS, there is no solution to it to the best of our knowledge.


\subsection{Compositional Semantic Parsing}

In contrast to neural semantic parsing models which are mostly constructed under a fully seq2seq paradigm, compositional semantic parsing models predict partial meaning representations and compose them to produce a full meaning representation in a bottom-up manner \cite{zelle1996learning-csp, zettlemoyer2012learning-csp, liang2013learning-csp, berant2013semantic-csp, berant2015imitation-csp, pasupat2015compositional-csp, herzig2020span-csp}.
Our model takes the advantage of compositional semantic parsing, without requiring any handcraft lexicon or syntactic rule.

\subsection{Unsupervised Parsing}



Unsupervised parsing (or grammar induction) trains syntax-dependent models to produce syntactic trees of natural language expressions without direct syntactic annotation \cite{klein2002natural-unsup,  bod2006all-unsup, ponvert2011simple-unsup, pate2016grammar, shen2018ordered-unsup, kim2019compound-unsup, drozdov2020unsupervised-unsup}.
Comparing to them, our model learns both syntax and semantics jointly.

\section{Conclusion}

In this paper, we introduce \textsc{LeAR}, a novel end-to-end neural model for compositional generalization in semantic parsing tasks.
Our contribution is 4-fold:
(1) \textsc{LeAR} focuses on algebraic recombination, thus it exhibits stronger compositional generalization ability than previous methods that focus on simpler lexical recombination.
(2) We model the semantic parsing task as a homomorphism between two partial algebras, thus encouraging algebraic recombination.
(3) We propose the model architecture of \textsc{LeAR}, which consists of a Composer (to learn latent syntax) and an Interpreter (to learn operation assignments).
(4) Experiments on two realistic and comprehensive compositional generalization benchmarks demonstrate the effectiveness of our model.

\section*{Acknowledgments} 

The work was supported by the National Key Research and Development Program of China (No. 2019YFB1704003), the National Nature Science Foundation of China (No. 71690231), Tsinghua BNRist and Beijing Key Laboratory of Industrial Bigdata System and Application.

\section*{Ethical Consideration}

The experiments in this paper are conducted on existing datasets. We describe the model architecture and training method in detail, and provide more explanations in the supplemental materials. All the data and code will be released with the paper. The resources required to reproduce the experiments is a Tesla P100 GPU, and for COGS benchmark even one CPU is sufficient. Since the compositional generalization ability explored in this paper is a fundamental problem of artificial intelligence and has not yet involved real applications, there are no social consequences or ethical issues.


\bibliographystyle{acl_natbib}
\bibliography{acl2021}

\clearpage
\appendix

\begin{table*}[h]
    \renewcommand\arraystretch{1.2}
    \centering
    \scalebox{0.85}{
    \begin{tabular}{cccc}
    \hline
    \textbf{Operation} & \textbf{Arguments} & \textbf{Result Type} & \textbf{Example} \\ \hline
    ON($t_1$, $t_2$) & \multirow{4}{*}{[$t_1$: Entity, $t_2$: Entity]} & \multirow{4}{*}{Entity} & Emma ate [\ub{the cake} on \ub{a table}] . \\ \cline{1-1} \cline{4-4} 
    IN($t_1$, $t_2$) &  &  & A girl was awarded [\ub{a cake} in \ub{a soup}] . \\ \cline{1-1} \cline{4-4} 
    BESIDE($t_1$, $t_2$) &  &  & Amelia dusted [\ub{the girl} beside \ub{a stage}] . \\ \cline{1-1} \cline{4-4} 
    \tabincell{c}{$\operatorname{ON}^{-1}$, $\operatorname{IN}^{-1}$, $\operatorname{BESIDE}^{-1}$} &  &  & NONE \\ \hline
    REC-THE($t_1$, $t_2$) & \multirow{4}{*}{[$t_1$: Entity, $t_2$: Entity]} & \multirow{4}{*}{Entity} & Lily gave [\ub{Emma} \ub{a strawberry}] . \\ \cline{1-1} \cline{4-4} 
    THE-REC($t_1$, $t_2$) &  &  & A girl offered [\ub{a rose} to \ub{Isabella}] . \\ \cline{1-1} \cline{4-4} 
    \tabincell{c}{AGE-THE, THE-AGE, \\ REC-AGE, AGE-REC } &  &  & - \\ \hline
    FillFrame($t_1$, $t_2$) & \tabincell{c}{[$t_1$: Entity, $t_2$: Pred/Prop] \\ {[}$t_1$: Pred/Prop, $t_2$: Entity]} & Proposition & A cat [\ub{disintegrated} \ub{a girl}] . \\ \hline
    CCOMP($t_1$, $t_2$) & \multirow{3}{*}{[$t_1$:  Pred/Prop, $t_2$:  Pred/Prop]} & \multirow{3}{*}{Proposition} & [\ub{Emma liked} that \ub{a girl saw}] . \\ \cline{1-1} \cline{4-4} 
    XCOMP($t_1$, $t_2$) &  &  & David [\ub{expected} to \ub{cook}] . \\ \cline{1-1} \cline{4-4} 
    $\operatorname{CCOMP}^{-1}$, $\operatorname{XCOMP}^{-1}$ &  &  & NONE \\ \hline
    \end{tabular}
    }
    \caption{
    Semantic operations in COGS. 
    ``Pred'' and ``Prop'' are abbreviations of ``Predicate'' and ``Proposition'', respectively.
    ``AGE'', ``THE'' and ``REC'' are abbreviations of ``AGENT'', ``THEME'' and ``RECIPIENT'', respectively.
    ``-'' omits similar examples.
    Some operations contain ``NONE'' example, indicating that no example utilize these operations in dataset.
    }
    \label{tab:COGS_operations}
\end{table*}

\begin{table*}[h]
    \renewcommand\arraystretch{1.2}
    \centering
    \scalebox{0.85}{
    \begin{tabular}{cccc}
    \hline
    \textbf{Operation} & \textbf{Arguments} & \textbf{Result Type} & \textbf{Example} \\ \hline
    UNION($t_1$, $t_2$) & \multirow{4}{*}{\tabincell{c}{[$t_1$: Entity/Prop, \\ $t_2$: Entity/Prop]}} & \multirow{4}{*}{Proposition} & what is the population of [\ub{var0} \ub{var1}] \\ \cline{1-1} \cline{4-4} 
    INTER($t_1$, $t_2$) &  &  & how many [\ub{cities named var0} \ub{in the usa}] \\ \cline{1-1} \cline{4-4} 
    \tabincell{c}{EXC($t_1$, $t_2$) \\ {EXC}$^{-1}$($t_1$, $t_2$) } &  &  & \tabincell{c}{which [\ub{capitals} are not \ub{major cities}]}
    \\ \hline
    \tabincell{c}{CONCAT($t_1$, $t_2$) \\ {CONCAT}$^{-1}$($t_1$, $t_2$)} & \multirow{1}{*}{[$t_1$: Pred, $t_2$: Pred]} & \multirow{1}{*}{Pred} & what is the [\ub{capital} \ub{of} var0] 
    \\ \hline
    FillIn($t_1$, $t_2$) & \tabincell{c}{[$t_1$: Entity/Prop, $t_2$: Pred] \\ {[}$t_1$: Pred, $t_2$: Entity/Prop]} & Proposition & how many [\ub{citizens} in \ub{var0}] \\ \hline
    \end{tabular}
    }
    \caption{
    Semantic operations in GEO. 
    ``Pred'' and ``Prop'' are abbreviations of ``Predicate'' and ``Proposition'', respectively.
    ``INTER'', ``EXC'' and ``CONCAT'' are abbreviations of ``INTERSECTION'', ``EXCLUDE'' and ``CONCATENATION'', respectively.
    }
    \label{tab:GEO_operations}
\end{table*}

This is the Appendix for the paper:
``Learning Algebraic Recombination for Compositional Generalization''.

\section{Semantic Operations in COGS}

The semantic primitives used in COGS benchmark are entities (e.g., \textit{Emma} and \textit{cat(x\_1)}), predicates (e.g., \textit{eat}) and propositions (e.g., \textit{eat.agent(x\_1, Emma)}).
The semantic operations in COGS are listed in Table \ref{tab:COGS_operations}.

The operations with ``$^{-1}$'' (e.g., ON$^{-1}$) are right-to-left operations (e.g., ON$^{-1}$(cake, table)$\rightarrow$table.ON.cake) while the operations without ``-1'' represent the left-to-right operations (e.g., ON(cake, table)$\rightarrow$cake.ON.table).
For operation FillFrame, the entity in its arguments will be filled into predicate/proposition as an AGENT, THEME or RECIPIENT, which is decided by model.

\An{\section{Semantic Operations in GEO and Post-process}}

\An{The semantic primitives used in GEO benchmark are entities (e.g., \textit{var0}), predicates (e.g., \textit{state()}) and propositions (e.g., \textit{state(var0)}).
The semantic operations in GEO are listed in Table \ref{tab:GEO_operations}.}

\An{To fit the FunQL formalism, we design two post-processing rules for the final semantics generated by the model.
First, if the final semantic is a predicate (not a proposition), it will be converted in to a proposition by filling the entity \textit{all}. 
Second, the predicate \textit{most} will be shifted forward two positions in the final semantics.}



\section{Policy Gradient and Differential Update}

In this section, we will show more details about the formulation of our RL training based on policy gradient and how to use differential update strategy on it.

Denoting $\tau = \{z, g\}$ as the trajectory of our model where $z$ and $g$ are actions (or called results) produced from Composer and Interpreter, respectively, and $R(\tau)$ as the reward of a trajectory $\tau$ (elaborated in Sec.~\ref{sec:reward}), 
the training objective of our model is to maximize the expectation of rewards as:
\begin{equation}
    \max_{\theta, \phi} \mathcal{J}(\theta, \phi)
    =\max_{\theta, \phi} \mathbb{E}_{\tau \sim {\pi}_{\theta, \phi }}\;R(\tau),
\end{equation}
where ${\pi}_{\theta, \phi }$ is the policy of the whole model $\theta$ and $\phi$ are the parameters in Composer and Interpreter, respectively.
Applying the likelihood ratio trick, $\theta$ and $\phi$ can be optimized by ascending the following gradient:
\begin{equation*}
    \nabla \mathcal{J}(\theta, \phi)=\mathbb{E}_{\tau \sim {\pi}_{\theta, \phi}}\;R(\tau)\nabla \log \pi_{\theta, \phi}\left(\tau\right),
\end{equation*}
which is same with Eq.~\ref{eq:gradient}.

As described in Sec.~\ref{sec:Model} that the interpreting process can be divided into two stages: interpreting lexical nodes and interpreting algebraic nodes, the action $g$ can also be split as the semantic primitives of lexical nodes $g_{l}$ and the semantic operations of algebraic nodes $g_{a}$.
In our implement, we utilize two independent neural modules for interpreting lexical nodes and interpreting algebraic nodes, with parameters $\phi_{l}$ and $\phi_{a}$ respectively.
Therefore, $\nabla \log \pi_{\theta, \phi}\left(\tau\right)$ in Eq.~\ref{eq:gradient} can be expanded via the chain rule as:
\begin{equation}\label{eq:gradient_split}
\begin{split}
    \nabla \log \pi_{\theta, \phi}\left(\tau\right) = 
    &\nabla \log \pi_{\theta}\left(z | x\right) + \\
    &\nabla \log \pi_{\phi_{l}}\left(g_{l} | x, z \right) + \\
    &\nabla \log \pi_{\phi_{a}}\left(g_{a} | x, z, g_{l} \right).
\end{split}
\end{equation}

With Eq.~\ref{eq:gradient_split}, we can set different learning rates:
\begin{equation}
\begin{split}
    &\theta \leftarrow \theta+\alpha\,{\cdot}\,{\mathbb{E}\;R(\tau)\nabla \log \pi_{\theta}\left(z | x \right)}, \\
    &\phi_{l} \leftarrow \phi_{l}+\beta\,{\cdot}\,{\mathbb{E}\;R(\tau)\nabla \log \pi_{\phi_{l}}\left(g_{l} | x, z \right)}, \\
    &\phi_{a} \leftarrow \phi_{a}+\gamma\,{\cdot}\,{\mathbb{E}\;R(\tau)\nabla \log \pi_{\phi_{a}}\left(g_{a} | x, z, g_{l} \right)}.
\end{split}
\end{equation}
Furthermore, in our experiments, the AdaDelta optimizer \cite{adadelta_2012-apd} is employed to optimize our model.

\section{Phrase Table}

\begin{table*}[t]
    \renewcommand\arraystretch{1.2}
    \centering
    \scalebox{0.85}{
    \begin{tabular}{ccc}
    \hline
    \textbf{Lexical Unit} & \textbf{Semantic Primitive(s)} & \textbf{Type} \\ \hline
    M0 & M0 & Entity \\ \hline
    \multirow{2}{*}{executive producer} & film.film.executive\_produced\_by & Predicate \\
     & film.producer.films\_executive\_produced & Predicate \\ \hline
    \multirow{3}{*}{editor} & a film.editor & Attribute \\
     & film.editor.film & Predicate \\
     & film.film.edited\_by & Predicate \\ \hline
    Italian & people.person.nationality m\_03rjj & Attribute \\ \hline
    \end{tabular}
    }
    \caption{Some examples in CFQ phrase table.}
    \label{tab:phrase table}
\end{table*}

The phrase table consists of lexical units (i.e., words and phrases) paired with semantic primitives that frequently co-occur with them.
It can be obtained with statistical methods.

For CFQ, we leverage GIZA++\footnote{https://github.com/moses-smt/giza-pp.git} \cite{och2003systematic-apd} toolkit to extract alignment pairs from training examples.
We obtain 109 lexical units, each of which is paired with 1.7 candidate semantic primitives on average.
Some examples in phrase table are shown in Table \ref{tab:phrase table}

As to COGS, for each possible lexical unit, we first filter out the semantic primitives that exactly co-occur with it, and delete lexical units with no semantic primitive.
Among the remaining lexical units, for those only contain one semantic primitive, we record their co-occurring semantic primitives as ready semantic primitives.
For lexical units with more than one semantic primitives, we delete the ready semantic primitives from their co-occurring semantic primitives.
Finally, we obtain 731 lexical units and each lexical unit is paired with just one semantic primitive.

\An{As GEO is quite small, we obtain its phrase table by handcraft.}


\begin{figure*}
  \centering
  
  \subfloat[][An example of generated results in CFQ benchmark with the input ``Did M6‘ s star, costume designer, and director influence M0, M1, M2, and M3 and influence M4 and M5 ''.\label{fig:cfq_tree}]
  {\includegraphics[width=1.\textwidth]{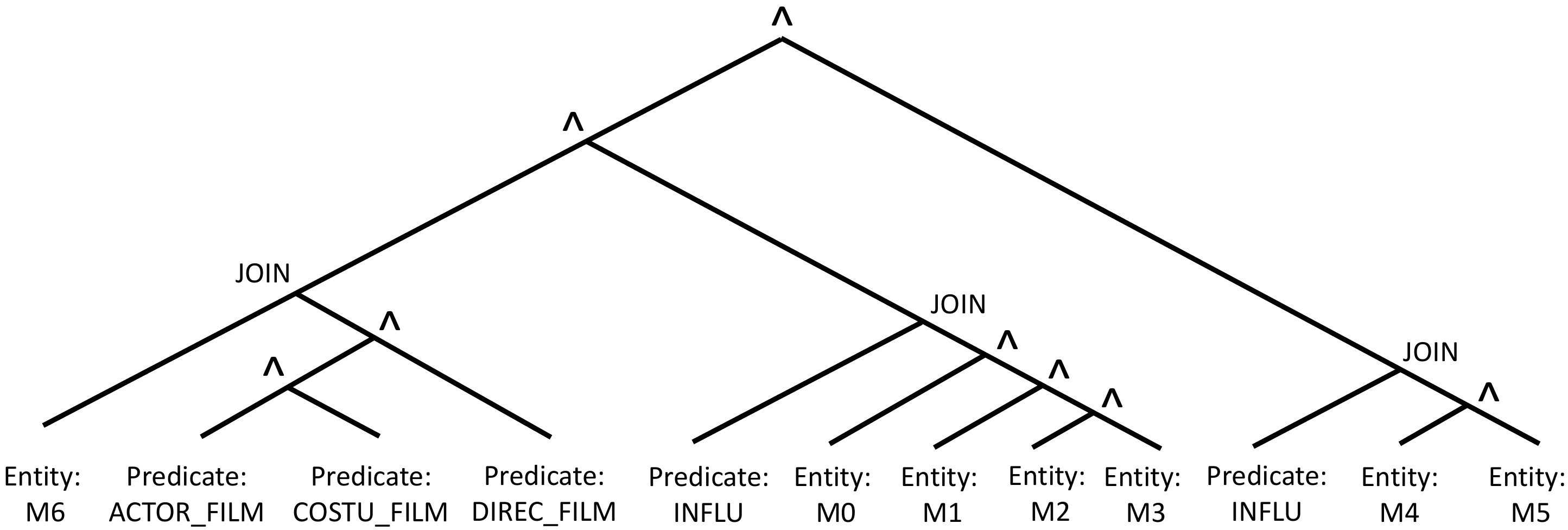}} \\
  
  \subfloat[][An example of generated results in COGS benchmark with the input ``Joshua liked that Mason hoped that Amelia awarded the hedgehog beside the stage in the tent to a cat''.\label{fig:cogs_tree}]
  {\includegraphics[width=1.\textwidth]{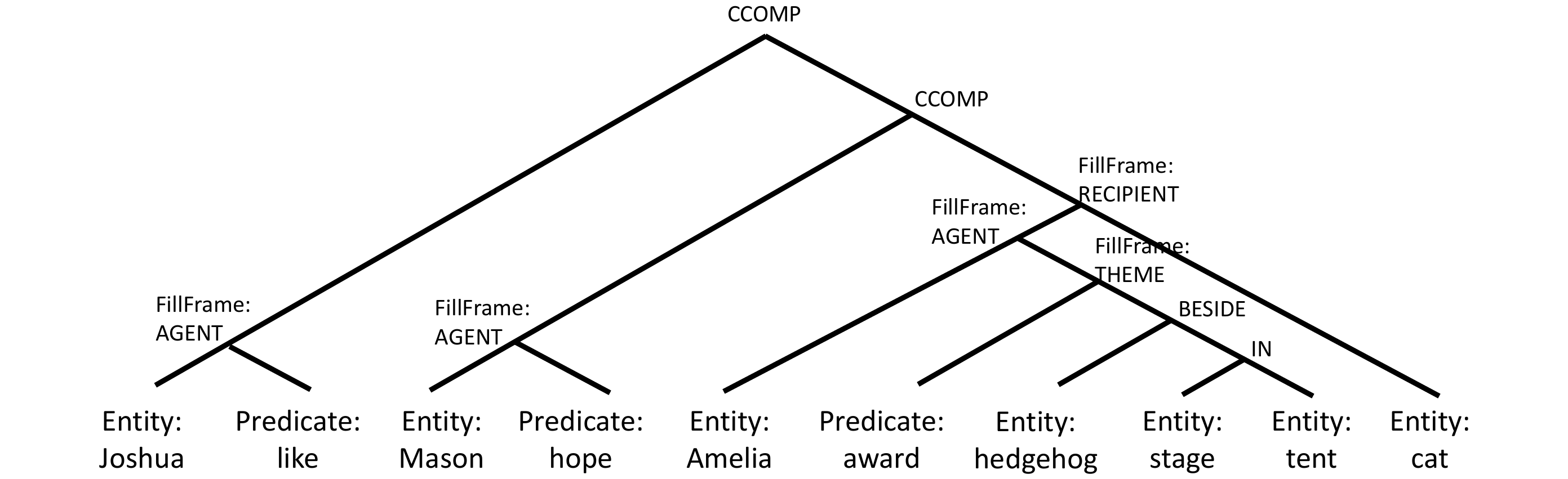}}
  
  \caption{Examples of generated tree-structures and semantics in CFQ and COGS benchmarks.
  }\label{fig:deep_trees}
  
\end{figure*}

\section{More Examples}

We show more examples of generated tree-structures and semantics in Figure~\ref{fig:deep_trees}.

\end{document}